\documentclass{article}
\usepackage{ijcai22}

\usepackage{times}
\usepackage{soul}
\usepackage{url}
\usepackage[hidelinks]{hyperref}
\usepackage[utf8]{inputenc}
\usepackage[small]{caption}
\usepackage{graphicx}
\usepackage{amsmath}
\usepackage{amsthm}
\usepackage{booktabs}
\usepackage{algorithm}
\usepackage{algorithmic}
\usepackage{amssymb}        

\usepackage{multirow}
\usepackage{booktabs}
\usepackage{tabularx}
\usepackage{makecell}
\usepackage{threeparttable}

\title{Mastering Asymmetrical Multiplayer Game with Multi-Agent Asymmetric-Evolution Reinforcement Learning}

\author{
	Chenglu Sun
	\and
	Yichi Zhang\and
	Yu Zhang\and
	Ziling Lu\and
	Jingbin Liu\and
	Sijia Xu\And
	Weidong Zhang
	\affiliations
	AI Lab, Netease\\
	\emails
	\{sunchenglu, zhangyichi1, zhangyu15, lzln6779, liujingbin01, xusijia, zhangweidong02\}@corp.netease.com
}

\begin{document}

\maketitle

\begin{abstract}
	Asymmetrical multiplayer (AMP) game is a popular game genre which involves multiple types of agents competing or collaborating with each other in the game. It is difficult to train powerful agents that can defeat top human players in AMP games by typical self-play training method because of unbalancing characteristics in their asymmetrical environments. We propose asymmetric-evolution training (AET), a novel multi-agent reinforcement learning framework that can train multiple kinds of agents simultaneously in AMP game. We designed adaptive data adjustment (ADA) and environment randomization (ER) to optimize the AET process. We tested our method in a complex AMP game named \textit{Tom \& Jerry}, and our AIs trained without using any human data can achieve a win rate of 98.5\% against top human players over 65 matches. The ablation experiments indicated that the proposed modules are beneficial to the framework.
\end{abstract}

\section{Introduction}
\label{Section1}
As a virtual simulation of reality, games are the good test platform for many AI researches. Recently, reinforcement learning (RL) has made rapid development in the field of game AI. Many AIs trained via RL methods have reached or even surpass top human level in different game genres, such as board games like \textit{Go} \cite{silver2018general}, card games like \textit{Mahjong} \cite{li2020suphx}, first-person shooting (FPS) games like \textit{Quake 3} \cite{Jaderberg2019human}, real-time strategy (RTS) games like \textit{StarCraft 2} \cite{vinyals2019grandmaster}, multiplayer online battle arena (MOBA) games like \textit{Dota 2} \cite{berner2019dota} and \textit{Honor of Kings} \cite{ye2020towards}. Those games have the characteristics of symmetry arena, and both sides in those games typically have the same initial resources, controlled rules, basic abilities, final goals, etc. These characteristics enable the self-play method to play a vital role in the RL training process. A large number of historical self-generated models obtained during the training are decent opponents with sufficient diversities and appropriate abilities. Battle against historical models iteratively can improve their capacities continuously and rapidly \cite{berner2019dota,vinyals2019grandmaster}, until got a powerful AI.

Asymmetrical multiplayer (AMP) game \cite{bycer2019asymmetrical,harry2021ten} is a popular game genre and a significant component in game field. The two sides in the AMP game usually have agents with different modes, such as \textit{Kick-and-Defend} \cite{bansal2018emergent}, \textit{Hide-and-Seek} \cite{baker2019emergent}, \textit{Tom \& Jerry}, \textit{Identity \uppercase\expandafter{\romannumeral5}}, \textit{Dead by Daylight}. Different modes of the two sides in the AMP game mean that the two sides are completely different in game rules, goals, number of players, etc. And the experience of playing one side cannot directly help play the other side\footnote{Although players in \textit{Dota 2} or \textit{StartCraft 2} can pick different heroes or races in different sides, they are still the symmetry games because both sides have the same rules, goals, etc. From another perspective, different heroes or races in symmetric games can be represented by different one-hot encoders or embeddings.}. Similar to the symmetrical games mentioned above, most AMP games are adversarial and zero-sum environments \cite{balduzzi2019open}. In these conditions, if we impose the self-play structure to train AMP games, one side usually has a higher win rate than the other side, leading to an imbalanced state distribution. That is, the state of one side is inclined to the advantage status, while the other side has more disadvantage states, which is not a comprehensive distribution. Besides, with further imbalances, it is hard for the weaker agent to get a positive reward, which brings a sparse reward issue \cite{trott2019keeping}. Hence, the AMP game is incapable of employing classical self-play training structure to enhance the AI performance. Previously, there is no powerful AI in the AMP game area that has achieved the level as the professional human players. \textit{Tom \& Jerry} is a typical 1v4 AMP game\footnote{\url{https://tomandjerry.fandom.com/wiki/Tom_and_Jerry_Chase}}, players can play as a cat or a mouse. When playing a cat, players need to fight alone against four mice, and when playing a mouse, players should cooperate with three other mouse players to fight against the cat. This game has the characteristics of various strategies, imperfect information, multi-agent cooperation, high-dimensional state space, and complex action operation, etc. Therefore, \textit{Tom \& Jerry} is an appropriate test bench for the AMP game area, Appendix A gives the details of its gameplay.

This paper proposes a multi-agent RL framework with the asymmetric-evolution training (AET) method. By using this framework, two modes of AI can be trained simultaneously for both sides of the game without using any human data. The mouse AI as a multi-agent model presents a superb level by defeating the top human players with a win rate of 98.5\%. During the AET process, we find that the characteristics of AMP game may cause the severe imbalance of win rate for both agents. This issue will result in a serious imbalance between mouse and cat agent levels, which further leads to stagnation of the AI performance. Some researchers \cite{bansal2018emergent,baker2019emergent} also found the imbalance issue in training AI for AMP games, while they focused on the evolvement and auto-curricula of the AI instead of addressing the imbalance phenomenon. To tackle the issue, we adopted adaptive data adjustment (ADA) and environment randomization (ER) to optimize the AET process. The main contributions of this work are as follows,

\begin{itemize}	
	\item A novel multi-agent deep RL framework with AET is proposed to train an AI that can defeat top players in a complex AMP game.
	\item Two modules are designed to solve the imbalance issue in AET, and experiments demonstrated that the designed modules are indispensable components.
\end{itemize}

\section{Related Works}
\label{Section:2}

The goal of this work is to train an AI with top human level in a large-scale AMP game with complex strategies as the unexplored research field. Previously, there have been many referable studies focused on training top-level AIs in the other different game genres.

In 2016, AlphaGo \cite{silver2016mastering} as a \textit{Go} AI defeated the world champion. AlphaGo was trained via RL combined with Monte Carlo Tree Search (MCTS) \cite{coulom2006efficient}. It utilized self-play method to continuously improve its performance, which demonstrated the powerful effects of self-play in the AI training of symmetric games. In 2018, a \textit{Dota 2} (a complex and adversarial MOBA game) AI called OpenAI Five defeated many top professional gamers \cite{berner2019dota}. OpenAI Five was trained using RL from scratch, its long self-play training process and huge history model pool guaranteed the diversity and stability of the AI. Its training process and reward reshaping offer guidance to our work. In 2019, AlphaStar \cite{vinyals2019grandmaster} as a \textit{StarCraft 2} AI achieved top human level. It used supervised learning (SL) to obtain a basic AI, and then utilized RL combined with league training frame to make the AI continually evolve. The elaborate network architecture for processing the complex entities and off-policy learning improvement method provide a reference for our work. In 2020, a \textit{Japanese Mahjong} AI called Suphx achieved 10 dan at Tenhou platform \cite{li2020suphx}. It also combined the SL with RL, and used global reward prediction with oracle guiding to improve the AI performance. In 2020, an \textit{Honor of Kings} AI called Juewu was introduced to defeat the top esports players utilizing RL with self-play training \cite{ye2020mastering,ye2020towards}. It successfully controlled most of the heroes with a designed ban-pick process. They improved the Proximal Policy Optimization (PPO) \cite{schulman2017proximal} via dual-clipping the advantage in the policy-gradient part, and we also used the dual-clip to maintain the stability of training. In addition to those above studies, many decent researches use RL to train multi-agent AI in the game field, such as \textit{Quake 3} \cite{Jaderberg2019human} and \textit{Google Research Football} \cite{kurach2020google}.

Those above games have the characteristics that the agents opposed in the game have the same mode. It means that their operation modes, restriction rules, and strategy ideas are almost the same. The two-side AIs can use the same training method, such as data processing and network. The difference between our work and the aforementioned works is that we simultaneously trained two AIs with totally different modes by the designed AET framework. Furthermore, the ADA and ER are combined to improve the framework and increase the diversity of AIs by addressing the imbalance issues. The easy-to-implement system design is conducive to the rapid iteration of the model. The experiments proved that the trained AI has reached a high level.

\section{Method}
\label{Section3}

\subsection{Asymmetric RL Training Framework}
\label{Section3.1}
\begin{figure*}[h]
	\centering
	\includegraphics[scale=0.32]{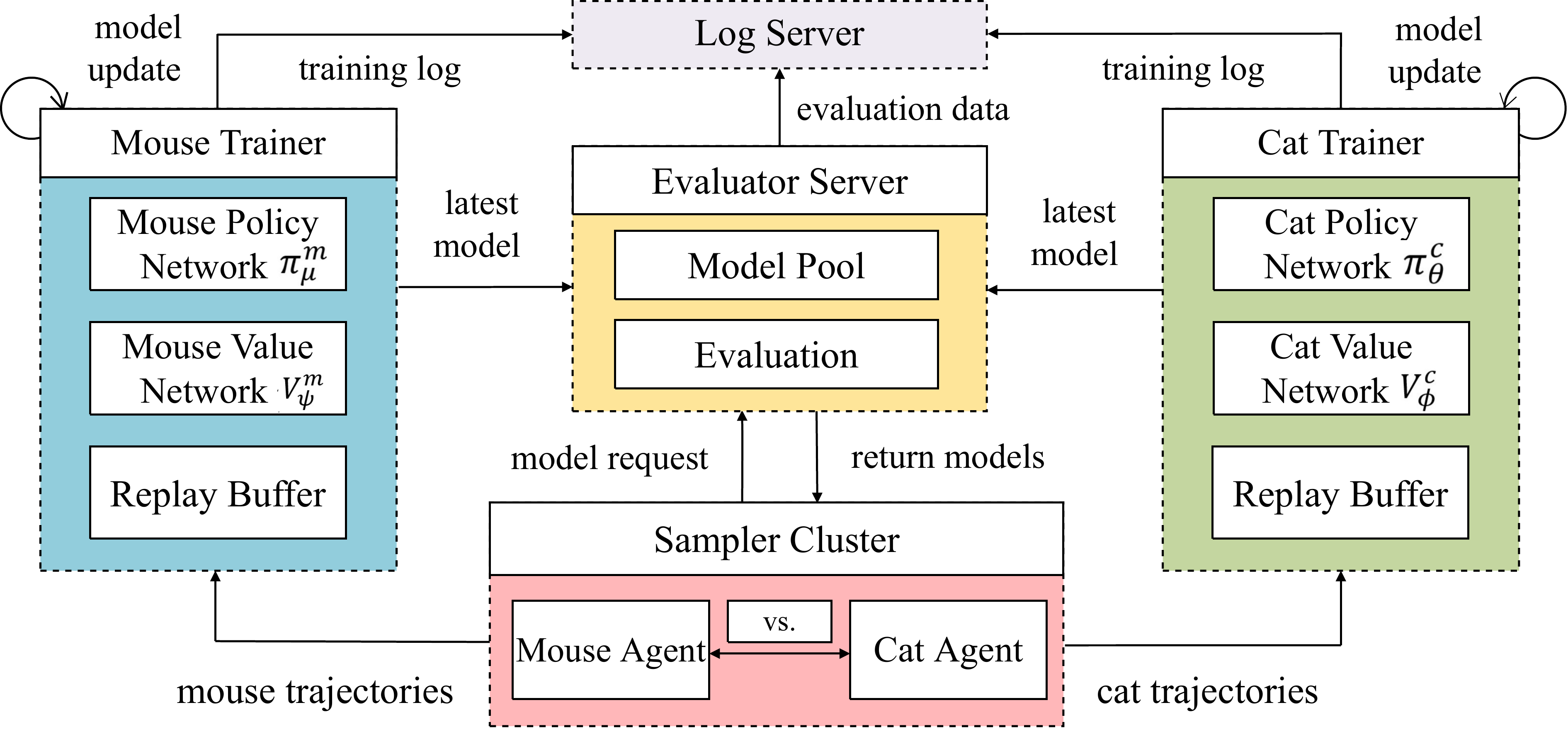} 
	\caption{Overview of the asymmetric RL training Framework.}.
	\label{fig:1}
\end{figure*}

The proposed training framework is mainly composed of five parts: mouse trainer, cat trainer, sampler cluster, evaluator server, and log server, as shown in Figure \ref{fig:1}. In the framework, a variant actor-critic structure is designed to train the agents. Different from the traditional actor-critic structure \cite{mnih2016asynchronous}, the variant simultaneously trains cat policy $ \pi^c_\theta(a^c_t|s^c_t) $, mouse policy $ \pi^m_\mu(a^m_t|s^m_t) $, cat value function $ V^c_\phi(\tilde{s}^c_t) $, and mouse value function $ V^m_\psi(\tilde{s}^m_t) $. $ s_t^c $ and $ s_t^m $ are the states with the information observed by the cat agent and mouse agent at timestep \textit{t}, respectively. Let $ \tilde{s} $ denote the state with invisible information. $ a_t^c $ and $ a_t^m $ are the actions selected by the cat and mouse agents, respectively. The $ \theta $ and $ \mu $ are the policy network parameters of cat agent and mouse agent, respectively. Similarly, $ \phi $ and $ \psi $ represent the parameters of value networks.

The trainer and sampler are built of GPU servers and CPU cluster, respectively. The mouse and cat agents built on their corresponding sampler cluster are competing in the game environment. The mouse trajectories extracted from game environments of the mouse samplers are sent to mouse replay buffers. And the cat trajectories have the similar transmission path. The trainers will periodically extract data from the replay buffers for model training and update the net parameters, followed by sending the latest model to the model pool. The evaluator server will perform the model evaluation on the CPU cluster. According to the results of model evaluation, cat and mouse agents will choose the appropriate model as opponents. Various logs in those modules are transmitted to the log server as feedback.

\subsection{Learning System}
\label{Section3.2}

\subsubsection{Algorithm}
\label{Section3.2.1}
Agents are trained using PPO \cite{schulman2017proximal}, a standard actor-critic algorithm \cite{konda1999actor}. The policy advantage is calculated via Generalized Advantage Estimation (GAE) \cite{schulman2015high}. Due to the asynchronous update of the models on the proposed distributed training framework, the dual-clipping PPO \cite{ye2020mastering} is used to mitigate instability in the off-policy training. The importance ratio of policy takes the form:
\begin{equation}
	\label{eq1}
	r^{k}_{t} = \frac{\pi^k(a^k_t|s^k_t)}{\pi^k_{old}(a^k_t|s^k_t)},  k \in {c,m}
\end{equation}
where $ \pi^k_{old} $ is the old policy of agent, and $ k $ represents cat or mouse agent.

The policy objective of the agent takes the form: 
\setlength{\arraycolsep}{0.0em}
\begin{equation}
	L^k_{p}=
	\begin{cases}
		-\mathbb{\hat E}_t[max(min(r^{k}_{t} {\hat A}^k_t, c^{k}_{t} {\hat A}^k_t), \eta^k{\hat A}^k_t],  & {\hat A}^k_t < 0\\
		-\mathbb{\hat E}_t[min(r^{k}_{t} {\hat A}^k_t, c^{k}_{t} {\hat A}^k_t)],  & {\hat A}^k_t \geq 0 \\
	\end{cases}
\end{equation}
where $ \mathbb{\hat E}_t[...] $ indicates the expectation over a finite batch of samples, $ c^{k}_{t} = clip(r^{k}_{t}, 1-\varepsilon^k, 1+\varepsilon^k) $ is the clipped ratio of PPO at timestep \textit{t}. $ {\hat A}^k_t $ is the advantage of the agent at timestep \textit{t}. $ \varepsilon^k $ and $ \eta^k $ are the original clip and the dual-clip PPO hyperparameters of agents, respectively, and $ k \in {c,m}  $.

The objective of the value network of the agent takes the form:
\begin{equation}
	\label{eq3}
	L^k_{v} = \mathbb{\hat E}_t[({V}^k(\tilde{s}^k_t) - G^k_t)^{2}],  k \in {c,m}
\end{equation}
where $ G^k_t = {V^k_{old}}(\tilde{s}^k_t) + {\hat A}_t^k $ is the returns of the agent, and $ V^k_{old} $ is calculated by the old value function. 

\subsubsection{Feature Processing}
\label{Section3.2.2}

The original data of the game can be acquired through the data interfaces provided by the game. Then the data are processed to the mini-images and vector information as the network input. The shape of the mini-image is 72 × 48 × 8, which has 8 channels, and 72 × 48 represents the reshaped size. The channel content of the mini-image is shown in Table \ref{table:1}. The locations of interactive entities are painted on the $ 2^{th}$ - $6^{th} $ channels by the order of importance of various entities.

\begin{table}[h]
	\centering
	\begin{tabular}{ll}
		\toprule
		Channels  &  Content                              \\
		\midrule
		$ 1^{th} $       & topographic map                       \\
		$ 2^{th}$ - $6^{th} $   & locations of interactive entities     \\
		$ 7^{th}  $      & location of cat agent                 \\
		$ 8^{th} $       & locations of mouse agents              \\
		\bottomrule
	\end{tabular}
	\caption{The channel content of the mini-image.}
	\label{table:1}
\end{table}

The vector information includes the content of the controlling agent, teammate, opponent, global status, interactive entities, and memory information, as shown in Table \ref{table:2}. The teammate information is only available for mouse agent, and the opponent information is the last seen information of the opponent. The main interactive entities include the cheeses, mouse holes, traps, items, etc. The memory information records many important historical messages, such as the historical actions and historical locations of various entities. 

\begin{table*}[h]
	\centering
	\begin{tabularx}{\textwidth}{ll}
		\toprule
		Types  &  Content                              \\
		\midrule
		Controlling agent       & \multirow{3}{13cm}{location, HP, character level, skill level, skill CD time, buff or debuff, holding items, etc.}        \\
		Teammate agent  &      \\
		Opponent agent      &                \\
		Global information       & game stage, time, progress, etc.              \\
		Interactive entities       & location, HP, distance to the players, main status, etc.              \\
		Memory information      & important historical information              \\
		\bottomrule
		\multicolumn{2}{l}{\small HP: health points; CD: cool-down.}\\
	\end{tabularx}
	\caption{The content of the vector information.}
	\label{table:2}
\end{table*}

\subsubsection{Network Architecture}
\label{Section3.2.3}
\begin{figure*}[h]
	\centering
	\includegraphics[scale=0.16]{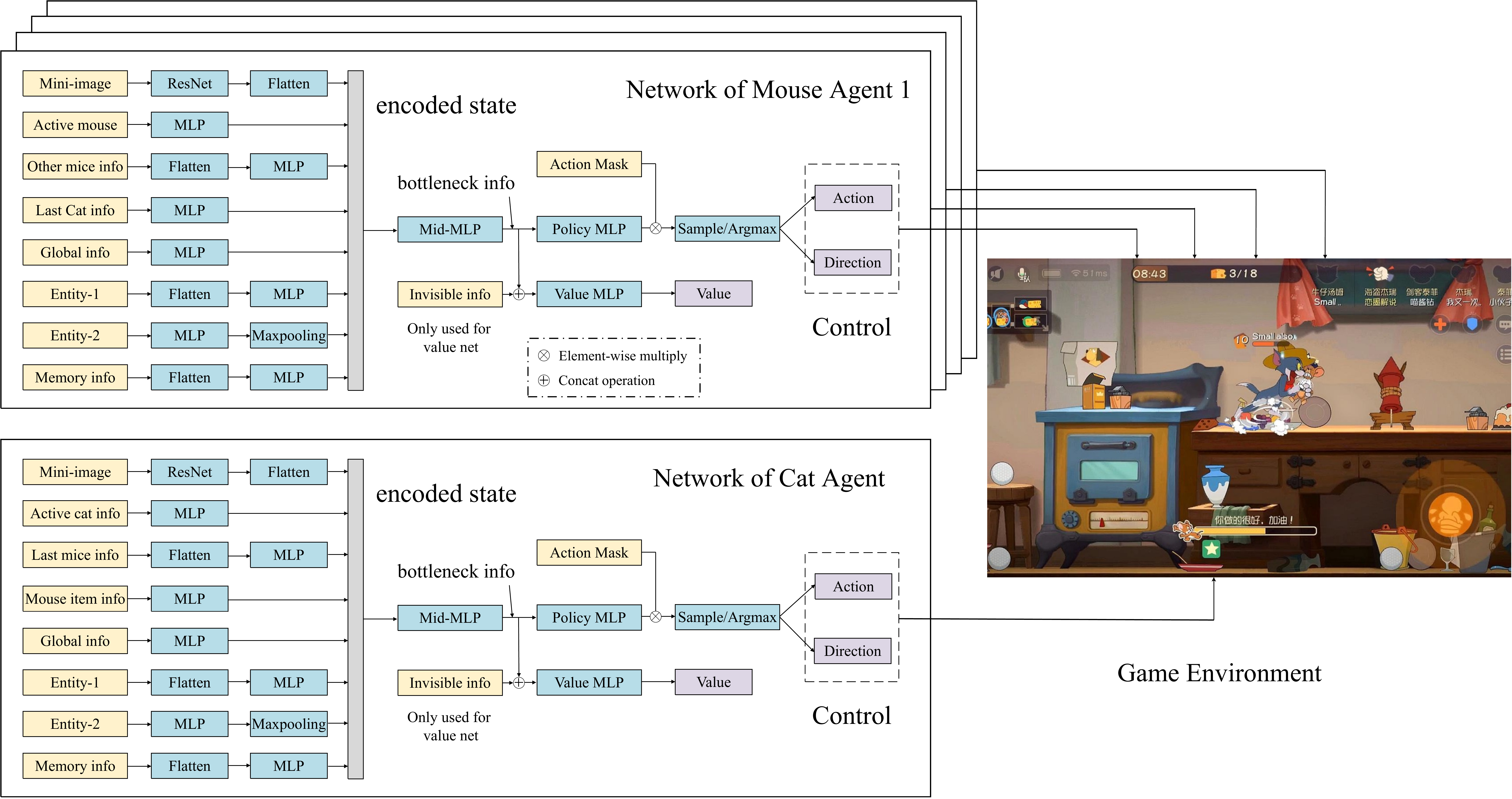}
	\caption{The overview of the network architecture. The inputs of the mouse agent and the cat agent are different. Active info summarizes the condition of the controlling agent. Other mice info represents most of the information that other mice can observe. Last info records the last status of opponents. Entity-1 represents the information of partial entities, and Entity-2 contains the information of entities with variable number. The Entity-2 is processed via an MLP and Max-pooling layer. The memory info will be embedded by one MLP. The output of the policy MLP will multiply with an action mask to get the action probability via a Softmax layer. The action mask is extracted based on the current state. Finally, the action and action directions will be selected by the sampling with probability distribution or argmax function. Relu is used in the Resnet and MLP as the activation function.}
	\label{fig:2}
\end{figure*}

The network mainly has three parts: state encoder, policy net, and value net. Figure \ref{fig:2} illustrates the overview of the network architecture. In the state encoder, the mini-image and the vector information are sent to the ResNet \cite{he2016deep} and multi-layer perceptron (MLP) for processing, respectively. The obtained hidden states are concatenated to acquire the encoded state, which is then transmitted to a Mid-MLP to get the bottleneck information. Finally, the bottleneck info is sent to the policy net and value net, which also adopt the MLP structure. Invisible information is added to train the value net for improving the value estimation. We conclude 18 actions and 16 action directions to control the agent's behavior, the action direction is used when casting some skills or items. The details of the network are provided in Appendix B. 

Our net structure did not use the Recurrent Neural Network (RNN) to store history information, because the embedding of memory information can make the model memorize important historical information without using RNN. Some previous explorations show that using memory information instead of Long short-term memory (LSTM) can speed up the training process and reduce the number of model parameters.

\subsubsection{Asymmetric Reward Shaping}
\label{Section3.2.4}

Reward will be given to the agent who completes some specific tasks in the game. Due to the characteristics of asymmetric game, two sets of reward shaping are required. In addition, the reward shaping process should assist to balance the win rate between the two different AIs for maintaining the well-proportioned history model pool. The shaped rewards have three different forms: ordinary reward, guidance reward, and team reward. Parts of guidance reward will gradually decay during the training process which is designed to prevent the agent from overfitting to some fixed strategies. Other guidance rewards are only awarded when the goal is achieved for the first time in each episode to prevent useless repeated actions. Team reward will be granted to the entire team after completing some important tasks (for mouse agents). The details of designed rewards are given in Appendix C. 

\subsection{Asymmetric-Evolution Training (AET)}
\label{Section3.3}
Self-play is an indispensable manner to improve the AI level of symmetrical battle arena game during the training process. By fighting against current self-model and historical self-models generated during the training process, the weakness of the training model can be quickly discovered. Defeat various models that have different strategies can make the AI have better generalization and performance.

\begin{figure}[h]
	\centering
	\includegraphics[scale=0.21]{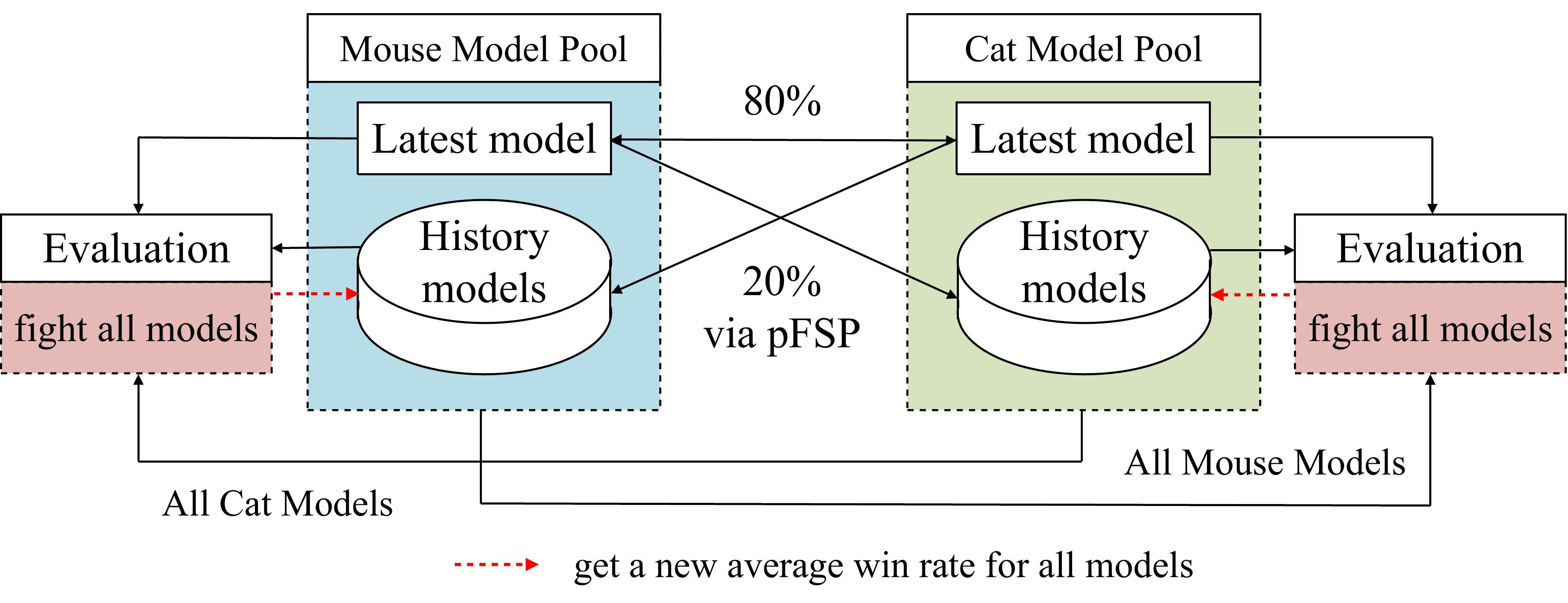}
	\caption{The overview of AET}
	\label{fig:3}
\end{figure}

For AMP games, the two sides are completely different AIs with different rules, goals, number of players, etc. The performance improvement of one side AI depends on the improvement of the overall level and diversity of both sides. Because overfitting phenomenon may occur if the AI performance of one side is stagnant. Hence, two different AIs in the AMP game must be trained and updated simultaneously to ensure that the AIs of both sides can be improved. Based on these above reasons, we propose the AET which simultaneously trains the cat and mouse agents as shown in Figure \ref{fig:3}. The two kinds of agents fight against the latest opponent models with 80\% probability, and compete historical models with 20\% probability. The latest models keep fighting against the historical models to avoid falling into a non-transitive situation \cite{gleave2019adversarial} and leaving strategic loopholes. We use pFSP \cite{vinyals2019grandmaster} to calculate the probability of choosing the history model from model pool, the probability of model \textit{a} fighting model \textit{b} is noted by

\begin{equation}
	\label{eq4}
	P^{b}_{a} = \frac{f(d^b_a)}{\sum_{c\subseteq C} f(d^c_a)}
\end{equation}
where $ d_a^b $ represents the win rate of model $ a $ defeating model $ b $, and model $ b $ is chosen from the opponent model pool $ C $. $ c $ represents the models in the opponent model pool $ C $. $ f(x)= (1-x)^p $ is the weighting function which control the importance of the win rate, $ p $ is set to 1 in this work. 

When a new model is added to the historical model pool (HMP), each model will fight all models in the opponent HMP for \textit{n} times as an evaluation process, to get a new average win rate for all models. The size of the HMP is 500 in this study, which is determined by the resource configuration. If the number of samplers for the agent is about 12,500, and 20\% of the samplers load the models in the HMP. By setting the HMP size as 500, each model can be loaded on about 5 samplers. Hence, with this configuration, the replay buffer can mostly have the data of all historical models, ensuring the diversity of opponents.

During the AET, the characteristics of AMP game will cause the imbalance of win rate for both agents. This phenomenon will result in unstable strategies and performance imbalance between two kinds of agents, yielding stagnation of both AI levels. The first reason is possible that, in the case of equal resources, one side’s higher win rate brings more “positive” trajectories per agent due to unbalanced game dynamics, but the other side has more “failure” trajectories, which will make the weaker side take more explorations to change the situation. However, the exploration usually brings more failures until finding a way to defeat the current opponent. The classical self-play method can avoid this problem. If the performance of the current model drops during training, the opponent’s latest model will also become weaker. Next, the current model will continually challenge the self-model and the fixed historical models to make up for the shortcomings. With further training and continuous exploration, the current model will finally defeat the model pool and have a better performance. But in the AET, the opponent will not stop improving and wait for the current model to become stronger. Secondly, when a side has better performance, its adverse situation is less in more cases. The stronger side will gradually forget how to deal with the disadvantaged situations, yielding poor performance in the adverse states. 

To address these two problems, we adopted adaptive data adjustment (ADA) and environment randomization (ER). In our environment, training mouse AI is much harder than training cat AI, because mouse AI is a multi-agent model, which requires complex cooperation to get a successful strategy. Furthermore, the win rate of mouse agents is almost zero in the early stage of training, because the mouse agents hardly complete tasks within a certain period via the random explorations. Hence, the mouse AI requires more resources at the early stage of training. But as the training progresses, the number of episodes experienced by the mouse agent is more than that of the cat agent, which leads to more exploration of the mouse. In this case, the mouse will gradually get out of its disadvantages, and the win rate of the cat will decline rapidly. Therefore, adopting a fixed ratio of sampler resources is hard to maintain a relative win rate balance. Hence, ADA is designed to dynamically adjust the sampler resources according to the evaluation results of the current model. For instance, when the win rate of the cat is larger than that of the mouse, the resources of the samplers will be tilted toward the mouse until the mouse AI gets a relatively balanced win rate with more exploration. The adaptive ratio of sampler resource of the mouse agent takes the form:
\begin{equation}
	\label{eq5}
	r^{m} = min(\alpha, max(\beta, r^m_{old} + dif))
\end{equation}
\begin{equation}
	dif = 
	\begin{cases}
		0.25(w^c - w^m),  & \text{if $w^c \geq w^m$} \\
		w^c - w^m,  & \text{otherwise} \\
	\end{cases}
\end{equation}
where $ \alpha $ and $ \beta $ are the upper limit and lower limit of the ratio, respectively. $ r^m_{old} $ indicates the old ratio of sampler resource of mouse agent. $ w^c $ and $ w^m $ are the win rate of cat and mouse agents, respectively. The values of $ \alpha $ and $ \beta $ depend on the imbalance characteristics of the two sides of the game, which are set to 0.8 and 0.5 in this work, respectively. Since the number of mouse agent is four times that of cat agent in each game, $ \alpha $ is set to 0.8 can keep the resources of mouse agent not exceeding four times the resources of cat agent. Setting $ \beta $ to 0.5 ensures that the hard-to-train side (mouse agent) have more resources than the easy-to-train side (cat agent). The new ratio of sampler resource $ r^m $ will be updated at a set interval. In practice, the samplers are controlled by the training framework with ADA and selectively train either the cat model or the mouse model. The framework will continuedly reallocate the resources until the target ratio is reached.

ER is used to expand the exploration cases, stable the training process, and balance the win rate of both sides by enabling the strong side to face more disadvantaged situations. When the unbalanced win rate between the two sides continues for a period, each episode of the game has a 45\% probability to increase the character level of the weaker side with a random value, and a 35\% probability to make the strong side face some random difficult cases, such as one teammate is eliminated. In addition, each episode has a 15\% probability that the weaker side will be strengthened for a while at the game beginning. These scenarios of ER are designed by analyzing the game data, observing the failure cases, and consulting with professional players. This method will continue to work until both sides have a balanced win rate. It can ensure that the strong side will get comprehensive training and handle various situations. With the exploitation of ADA and ER, both sides can dynamically maintain a relatively balanced situation, consequently achieving the goal of continuous evolution.

\section{Experiments}
\label{Section:4}

The whole training process costs 138 Nvidia 2080Ti GPUs and 55,000 CPUs. The parameter numbers of the cat model and mouse model are 4.37 million and 4.89 million, respectively. The resource allocation between mouse agent and cat agent is adjusted in real-time as introduced in the last section. 

\subsection{Results of Matches Against Human Players}
\label{Section:4.1}
The AI model is deployed on game server, and each game client sends the action request to the server. The communication between the client and the server is asynchronous, i.e., the client will execute the last action if not receive the new action from the server at the current frame. The frames per second (FPS) of the game is 15, and the time-delay is set to 3 frames. The client requests the action every 2 frames, and the intermediate frame will continue to execute the last action. Hence, the average response time is (3 + 4) / 2 / 15 = 0.233s. Due to the network communication fluctuation, the practical average response time is 0.252s by deployment testing.

There are six ranks in the game: Bronze, Silver, Gold, Platinum, Diamond, and Master, respectively representing the top 100\%, 70\%, 45\%, 20\%, 5\% and 1\% players. Twenty players were invited to challenge our AI, which included 5 Master players, 8 Diamond players, and 7 Platinum players. The 5 Master players are game reviewers, and one of them has won a championship with Top 1 ranking. To prevent the influence of temporary teamwork of human players, we make those human players control the cat to play against our mouse AI. Human players can choose one cat character from \textit{Tom}, \textit{Lightning}, \textit{Bodyguard Tom}, \textit{Tara}, and \textit{Fencer Tom}. A total of 65 games were played, and our mouse AI won 64 games with a win rate of 98.5\%. Our AI played 20 games with professional players, and won 19 of them. Those professional players got an average score of 1619.2 per episode, 4.05 mice caught and 0.65 mice eliminated per episode. The AI won left 45 games against the other 15 players, and those players got an average score of 856.6 per episode, 1.62 mice caught and 0.11 mice eliminated per episode. Through the testing, our AI demonstrated a robust performance in different adverse situations, especially playing against those professional players. The professional players believed that the AI has the top level in Master rank and has various strategies and excellent human-like behavior. Most of game episodes played between our agents and top human players are available at: \url{https://sourl.cn/fjhGMG}, in which the observation view of human player is employed.

\subsection{Training Process}
\label{Section:4.2}

\begin{figure}[h]
	\centering
	\includegraphics[scale=0.29]{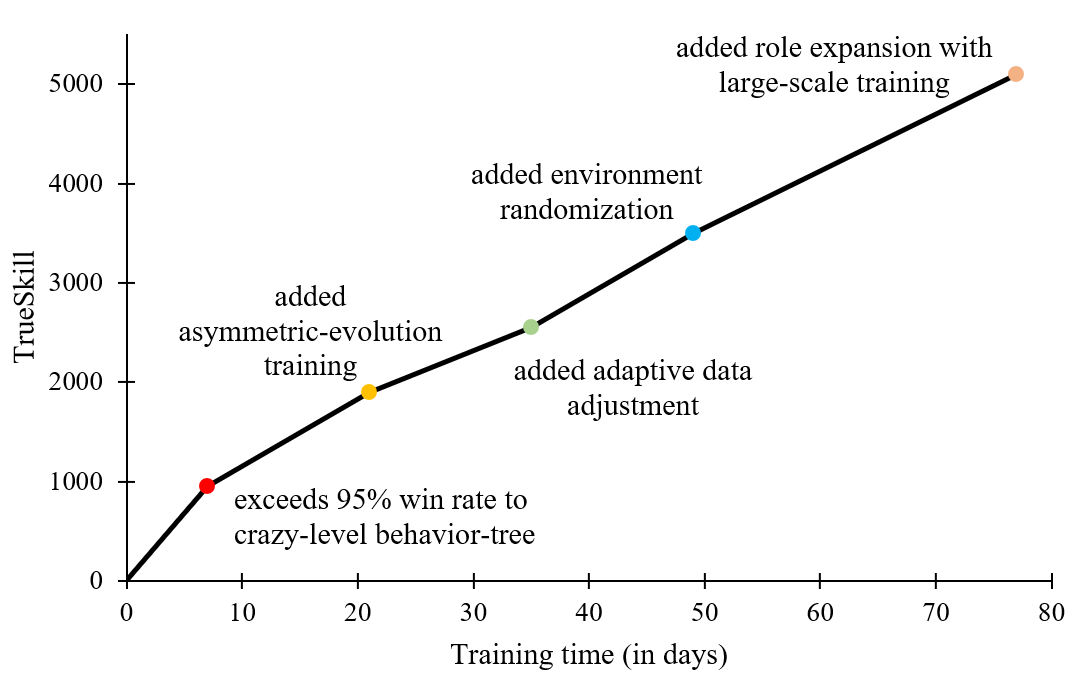}
	\caption{The training process.}
	\label{fig:4}
\end{figure}

The model was first trained to defeat the crazy-level bot to obtain a basic model. Crazy-level bot is the highest-level build-in AI in the game, which is elaborately designed with the behavior trees \cite{iovino2022survey} by the game studio. The rank of the crazy-level bot is around the junior Gold level reported by the game studio. The training process takes 7 days to get the basic model from scratch, which can have a win rate of more than 95\% to the crazy-level bot. Various branch versions of training were conducted by fine-tuning the basic model. By exploration, the joins of AET, ADA, ER, and role expansion with large-scale\footnote{The ``large-scale'' here refers to the utilization of more GPU, CPU resources and larger related parameters such as batch size.} training are important improvement nodes. The role expansion with large-scale training is the final stage that added more opponent characters and training resources. The training process of the proposed AI is shown in Figure \ref{fig:4}. As mentioned above, the final version of mouse AI can achieve a win rate of 98.5\% against human players, and the final version of the cat AI can have a win rate of more than 40\% against the final mouse AI. We use the Trueskill score \cite{herbrich2006trueskill} as an indicator to evaluate the model performance. Assume that the crazy-level bots of cat and mouse have the same TrueSkill score. The mouse AI was trained with a fixed lineup including \textit{Detective Jerry}, \textit{Robin Hood Jerry}, \textit{King Jerry}, and \textit{Tuffy}. The training hyperparameters are provided in Appendix B, which were selected through a series of search tests in the early stage of the project. Due to the huge combination of hyperparameters, we cannot guarantee that this set of hyperparameters is the optimal choice, but it can guarantee good results for the similar applications. The training process were implemented by Python 3.6 with Pytorch 1.6.

\begin{figure*}[htbp]
	\centering
	\includegraphics[scale=1.1]{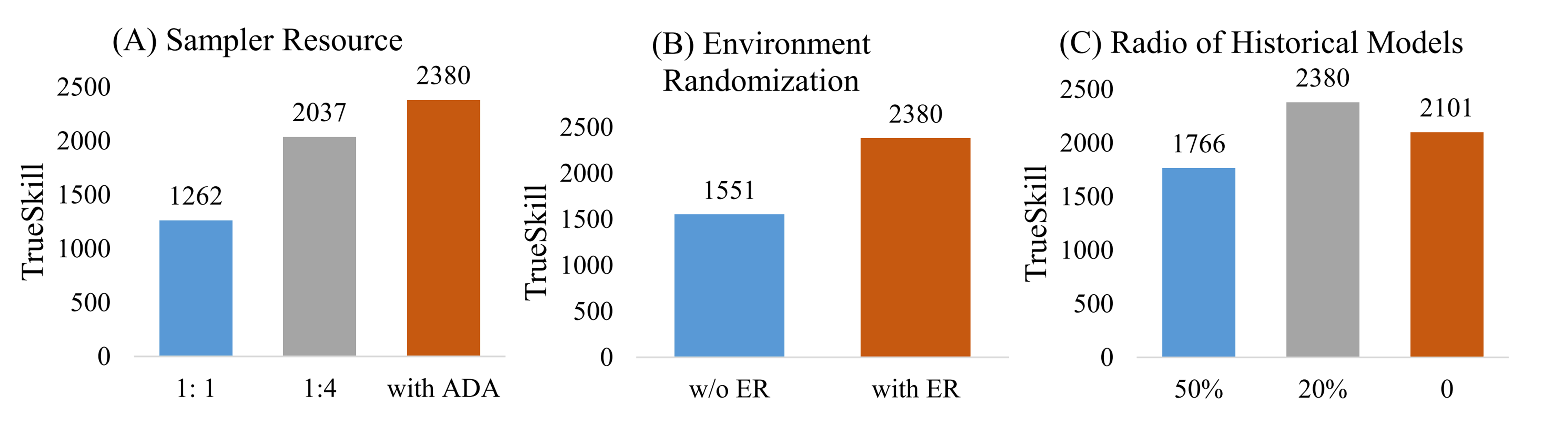}
	\caption{The TrueSkill results of ablation tests for key components in the AET. (A) Comparing different allocations of sampler resource between cat and mouse agent (1: 1, 1: 4, use ADA); (B) Comparing the effects with/without ER; (C) Comparing different ratios of the historical models (50\%, 20\%, 0).}
	\label{fig:5}
\end{figure*}

\begin{figure}[htbp]
	\centering
	\includegraphics[scale=0.21]{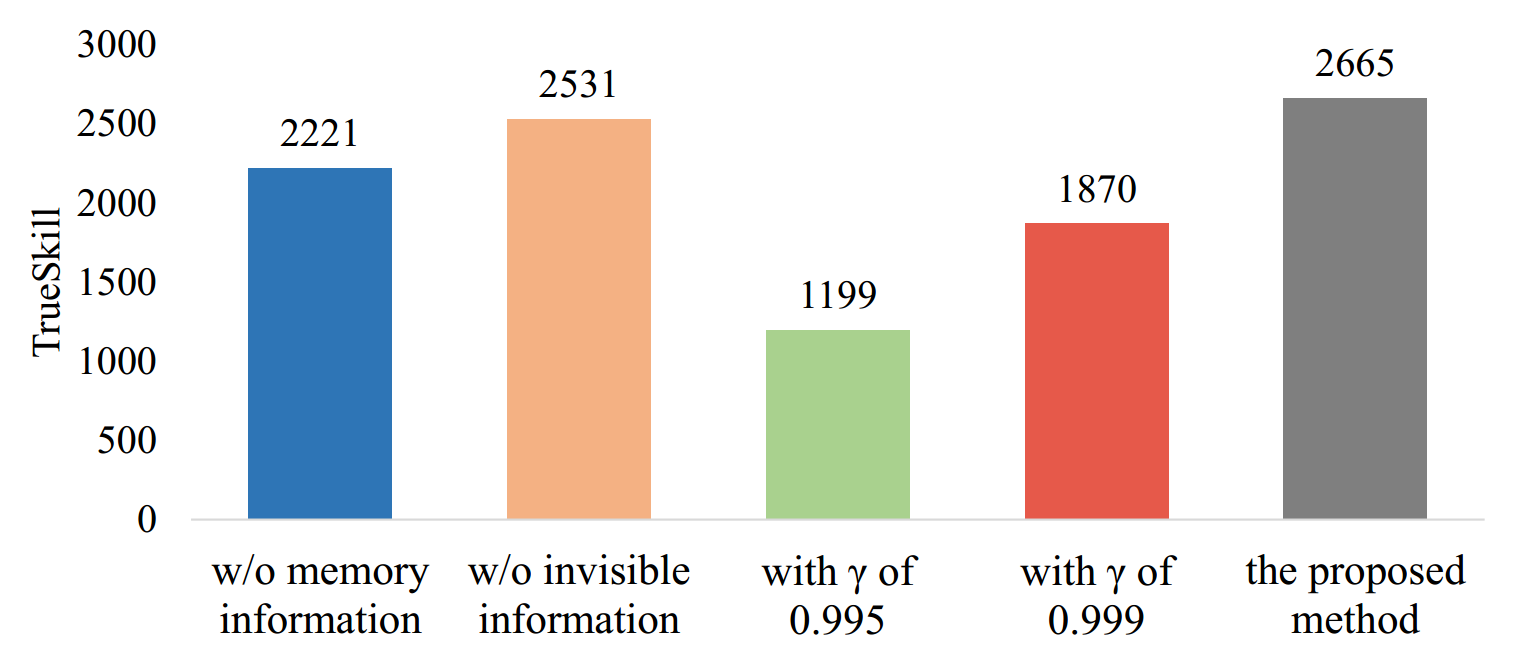}
	\caption{The TrueSkill results of different factors adopted in the learning system.}
	\label{fig:6}
\end{figure}

\subsection{Ablations}
\label{Section:4.3}

Two groups of ablation experiments were conducted to identify the effect of key components in the proposed method. As mentioned in Section 3.3, the ADA and ER are two key components in the AET. Besides, the scale of historical models in the training process is also important. Hence, the first group of experiments conducted three ablation tests to validate the effectiveness of those components. The basic models of mouse and cat agent as the converged models trained against the crazy-level bots were used as the starting models for the test. The Trueskill value of the mouse AI after 7-day training was used as the evaluation indicator. Figure \ref{fig:5} (A) demonstrates that training with ADA is beneficial to the AI performance, while keeping the resource balance is the worst choice for our asymmetrical environment. In other words, the successful experience of training process in those symmetry environments cannot be simply adopted in the AMP environments. The ER can significantly increase learning effectiveness as shown in Figure \ref{fig:5} (B). It shows that increasing the diversity of samples, especially the samples of difficult scenes, is conducive to improve the performance and generalization of the model. Figure \ref{fig:5} (C) shows that training with 20\% of historical models leads to a better performance than that using 50\% of historical models or without using historical models. The reason maybe that a large proportion of fighting the historical models with fixed weights could make the training model overfit to some fixed strategies to easily obtain the positive rewards, which would also slow down the improvement of the model. And an appropriate proportion of fighting the history models can improve the model performance by increasing the diversity of samples. From the experimental results of the first group, it can be seen that the traditional self-play settings (without ADA and ER) cannot work well for the asymmetric environments.

In the second group, we compared three factors adopted in the learning system, including memory information,  invisible information for the value net, and the selection of discount factor $ \gamma $. Five mouse AIs with different settings were trained against the cat models for 7 days, and the cat models were trained using the proposed method in this paper. Figure \ref{fig:6} presents the Trueskill scores between those different mouse AI versions. The result shows that using memory information can increase the training effect. The invisible information is conducive to the training by improving the value estimation. The parameter $ \gamma $ is another important factor that using the value of 0.995 and 0.999 has much lower Trueskill scores than the proposed method whose $ \gamma $ is set to 0.9995. It shows that the actions may have a long-term impact to the strategy in this game. The training process of each test costs 4000 samplers in those ablation experiments.

\section{Conclusion, Limitations \& Future Work}
\label{Section:5}
This paper proposes a multi-agent deep RL framework with asymmetric-evolution training method. The adaptive data adjustment and environment randomization are designed to solve the imbalance issues in the training process. We show that, for the first time, RL-based agents can achieve top human-level performance in a complex asymmetrical multiplayer game without using any human data. The ablation experiments proved that the designed modules are beneficial to the training process. However, our work still has some limitations. Similar to other AIs that have reached the level of top human players, the training process injects human expertise related to the environment, such as reward shaping. In the future, we attempt to construct intrinsic rewards \cite{zheng2020can} to replace the complicated reward shaping process. Besides, \textit{Tom \& Jerry} has many roles, but we only trained parts of roles due to the expensive computational cost. Hence, in the future, we will make the trained AI support full roles in this game and hold a competition that challenges the top human players with the cat AI. In addition, the hyperparameters used in the training process are probably not the best choice on other environments, so the adjustments are still required for a new study. Currently, there is no open source asymmetric multiplayer game environment with complex and various strategies, so our study is tested on a proprietary game environment. Next, we will try to develop an open source asymmetric multiplayer test bench and make our asymmetrical training framework open source to train more AIs.

\clearpage
\bibliographystyle{named}
\bibliography{ijcai22}

\appendix

\section{Introduction of the Game \textit{Tom \& Jerry}}
\label{appendix:1}

\textit{Tom \& Jerry} is a typical 1v4 asymmetrical multiplayer online game. The two sides in the game have different resources, rules, basic abilities and final goals. Players will play in a scene which is composed of utility room, dining room, living room, bedroom, yard, kitchen, attic, and other rooms. Four mouse players and one cat player will carry out a series of complex interactions for completing their mission in the scene. At the start of the game, the game scene will be generated in which the positions of those rooms are random. In those rooms, five mouse holes and five pieces of cheese will be randomly generated. The mouse team needs to cooperate to push five pieces of cheese into those mouse holes in the scene and then break through a wall crack to escape. Each mouse hole can only be stuffed with one cheese. One wall crack as the escape way will be randomly generated somewhere in the scene after all the cheeses have been pushed into the mouse holes. Mouse victory is achieved by two or more mice escaping. During the game, the mice are possible to be caught by the cat and tied to a rocket. If no teammates come to rescue within a certain period, the mouse tied to a rocket will be eliminated. The cat fights alone to prevent the mice from obtaining the cheeses and escaping. The cat will win if successfully eliminates 3 mice, or prevents the mice from pushing in 5 pieces of cheese within a specified time, or prevents 3 mice from escaping.

\begin{figure*}[h]
	\centering
	\includegraphics[scale=0.2]{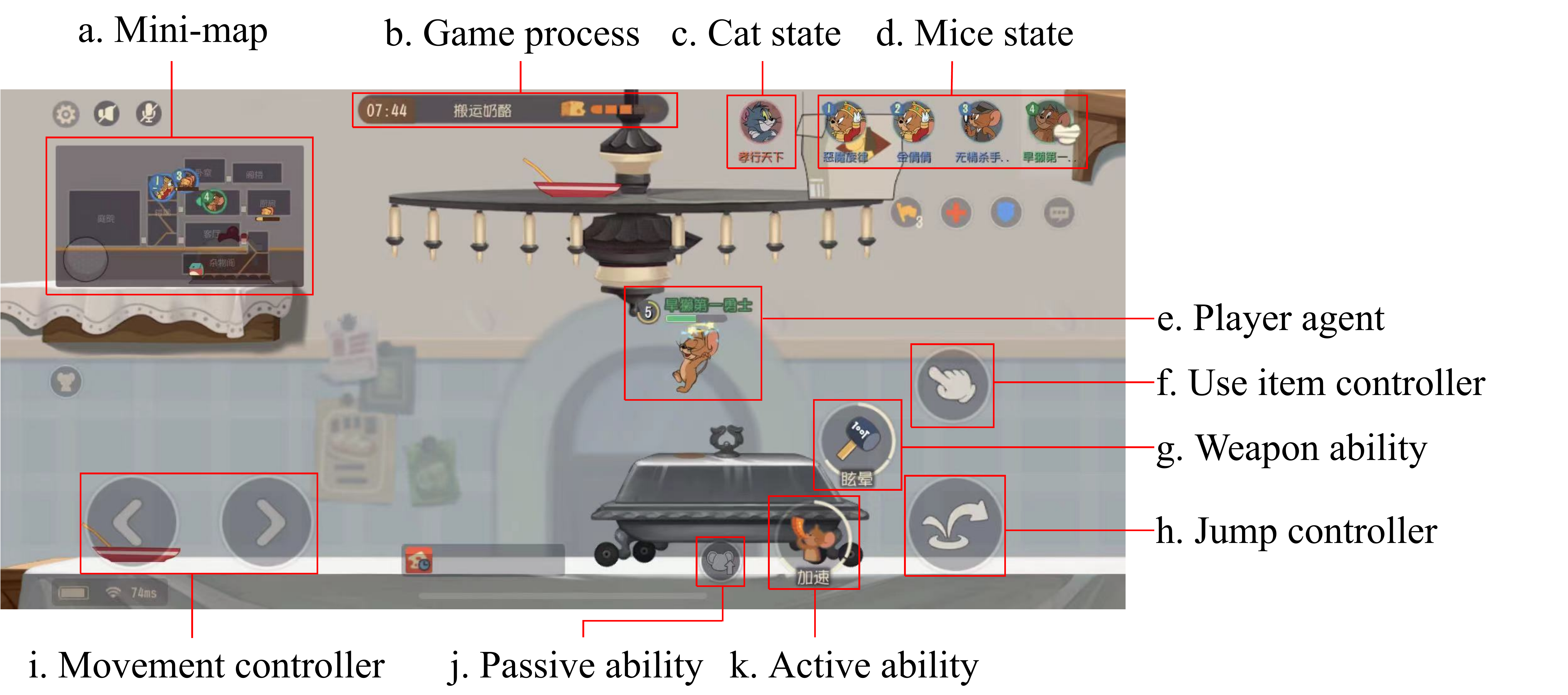}
	\caption{Game UI of \textit{Tom \& Jerry}. (a)-(d) present the global game information, (e) is the mouse agent controlled by the player, (f)-(k) are control buttons.}
	\label{fig:7}
\end{figure*}

The mouse agent or cat agent has one active ability, one passive ability, and one or two weapon abilities. Those abilities have different functions, which are helpful to complete the final goal. There are a large number of different types of items that can be acquired and used in the game. In addition, the non-player characters (NPC) will also influence the game progress, for example, the dog in the scene will attack nearby players indiscriminately. Since almost all important issues (cheeses, mouse holes, items, etc.) in the scene are randomly generated, the game has a short preparation stage. In the preparation stage, each mouse player can control a scout vehicle to observe the scene and decide the team strategy according to the layout of the scene, and the cat can seek and destroy the scout vehicle to improve the role level. After the preparation phase is over, the game starts. Those features greatly increase the complexity, playability and interest of the game. In Figure \ref{fig:7}, the user interface (UI) of the \textit{Tom \& Jerry} is given as an example. In this environment, the action space is \textasciitilde10e9712 (144 discretized actions, 4,500 frames per game). The state space is \textasciitilde10e1769 (3,456 discretized positions, 5 agents, 100+ state each agent). \textit{Tom \& Jerry} run at 15 FPS for more than 5 minutes, so the episode length is more than 4,500 steps, and the agents will get one reward every 150 steps on average.

\section{Details of the Network Architecture and Training Hyperparameters}
\label{appendix:2}

Table \ref{table:3} lists the details of the designed network architecture. (18, 16) in the policy-MLP represents that it has two policy heads which includes the actions and action directions. Table \ref{table:4} lists the details of the training hyperparameters. The sample reuse is a floating value, because there are communication fluctuations in the training framework, which could cause the changing of data transmission time during the training process.

\begin{table}[h]
	\centering
	\begin{tabular}{ll}
		\toprule
		Network structure  &  Size                              \\
		\midrule
		ResNet       & \makecell[l]{3 Res-blocks, each output \\ channel is 32, 64, 128}                       \\
		MLP in the state encoder   & 256 × 128     \\
		Mid-MLP     & 1280 × 1024                \\
		Policy-MLP      & 1024 × 512 × 256 × (18, 16)              \\
		Value-MLP       & 1024 × 512 × 256 × 1              \\
		\bottomrule
	\end{tabular}
	\caption{The details of the network architecture.}
	\label{table:3}
\end{table}

\begin{table}[h]
	\centering
	\begin{tabular}{ll}
		\toprule
		Hyperparameter     & Value  \\
		\midrule
		Frame skip & 2    \\
		Batch size     & 7000 per GPU      \\
		Trajectory length     & 128     \\
		Sample reuse     & 1.0 $\sim$ 1.2     \\
		PPO clipping     & 0.2     \\
		PPO dual-clipping     & 3     \\
		Discount factor $ \gamma $     & 0.9995      \\
		GAE discount $ \lambda  $    & 0.95      \\
		Value loss weight   & 0.5     \\
		Entropy coefficient   & 0.01     \\
		Optimizer   & Adam     \\
		Learning rate     & 5e-5      \\	
		Adam $ \beta_1, \beta_2 $    & 0.99, 0.999      \\	
		\bottomrule
	\end{tabular}
	\caption{Training hyperparameters.}
	\label{table:4}
\end{table}

\section{Details of the Reward Shaping}
\label{appendix:3}

Table \ref{table:5} and Table \ref{table:6} lists the rewards of the cat agent and the mouse agent, respectively. Those rewards are designed to encourage the agents to complete the important tasks. The guidance rewards are used to prevent the agents from ``diddling'' the reward through easy tasks.

\begin{table*}[h]
	\centering
	\begin{tabular}{lcc}
		\toprule
		Tasks  &  Value   &  Guidance                         \\
		\midrule
		Place the mouse-trap   & 0.025   &                    \\
		Catch the mouse   & 0.1 &    \\
		Tie the mouse to the rocket     & 0.1  &              \\
		Eliminate one mouse      & 2   &           \\
		Cheese pushed into mouse hole       & -0.25       &       \\
		HP change of the wall crack       & 0.01 × CV       &       \\
		Win / Lose       & 5 / -5       &       \\
		Find rockets for the first time       & 0.05       &   $ \surd $    \\
		Find gold hole for the first time       & 0.05       &   $ \surd $    \\
		Damage mouse for the first time       & 0.15       &   $ \surd $    \\
		Destroy a scout vehicle (preparation stage)       & 0.5       &       \\
		Lose / Recapture cakes (preparation stage)       & -1 / 1       &       \\
		\bottomrule
		\multicolumn{2}{l}{\small HP: health points; CV: change value.}
	\end{tabular}
	\caption{Reward list of the cat agent.}
	\label{table:5}
\end{table*}

\begin{table*}[h]
	\centering
	\begin{threeparttable}
		\begin{tabular}{lccc}
			\toprule
			Tasks  &  Value   &  Guidance   &   Team                    \\
			\midrule
			Distance to the nearest cheese   & 0.001 × DI × DC (from 5 to 1)   &   $ \surd $   &               \\
			\makecell[l]{Distance to the nearest mouse who is tied to rocket}   & 0.01 × DI × DC (from 3 to 1) &  $ \surd $   & \\
			\makecell[l]{Distance to the nearest mouse hole while lifting the cheese}     & 0.001 × DI × DC (from 5 to 1)  &   $ \surd $    &        \\
			Distance to the wall crack      & 0.001 × DI × DC (from 5 to 1)   &  $ \surd $    &      \\
			Pick up the cheese       & 0.05 × Coef (from 1 to 0)       &  $ \surd $  &    \\
			HP change       & 0.003 × CV       &     &   \\
			Caught by the cat       &  -0.2       &      &  \\
			Eliminated by the rocket       & -2       &      &  \\
			Save teammate from the cat or the rocket       & 0.5       &     &  \\
			\makecell[l]{Throw the explosive when the explosive is about to explode}       & 0.15       &       & \\
			Destroy the gloves at key locations       & 0.15      &      &  \\
			Destroy the mouse traps at key locations       & 0.15       &       & \\
			Pushing the cheese       & 1 × progress / NMPC       &       & \\
			Push the cheese into the mouse hole       & 0.5       &       & $ \surd $ \\
			Damage the wall crack      & 2 × DP       &       & \\
			Open the wall crack       & 1       &       & $ \surd $ \\
			Win / Lose       & 5 / -5       &       & $ \surd $ \\
			Scout vehicle destroyed (preparation stage)       & -0.5       &       & \\
			Lose / Recapture cakes (preparation stage)       & 1 / -1       &       & \\
			\bottomrule
		\end{tabular}
		\begin{tablenotes}
			\footnotesize
			\item DI: Distance; DC: Decay coefficient; Coef: coefficient; NMPC: the number of mice pushing cheese; DP: damage percent. 
		\end{tablenotes}
	\end{threeparttable}
	\caption{Reward list of the mouse agent. Parts of guidance reward will gradually decay in three days, and the team reward will be given to all the mouse agents when complete the corresponding tasks.}
	\label{table:6}
\end{table*}

\end{document}